\documentclass{article}

\usepackage{arxiv}

\usepackage[utf8]{inputenc} 
\usepackage[T1]{fontenc}    
\usepackage{hyperref}       
\usepackage{url}            
\usepackage{booktabs}       
\usepackage{amsfonts}       
\usepackage{nicefrac}       
\usepackage{microtype}      
\usepackage{lipsum}
\usepackage{graphicx}
\graphicspath{ {./images/} }
\usepackage{amsmath}

\usepackage{graphicx}
\usepackage{comment}
\usepackage{amsmath,amssymb} 
\usepackage{color}
\usepackage{xfrac}

\usepackage{enumitem}   
\usepackage{floatrow}
\newfloatcommand{capbtabbox}{table}[][\FBwidth]

\usepackage{blindtext}

\usepackage{floatrow}

\title{End-to-end Learning Improves Static Object Geo-localization in Video}

\author{
     Mohamed Chaabane$^{1,2}$, Lionel Gueguen$^{2}$, Ameni Trabelsi$^{1}$, Ross Beveridge$^{1}$, and Stephen O'Hara$^{2}$  \\
\\
 $^1$ Colorado State University, Fort Collins CO 80523, USA\\
 $^2$ Uber Advanced Technologies Group, Louisville CO 80027\\
}

\begin{document}
\maketitle

\begin{abstract}
Accurately estimating the position of static objects, such as traffic lights, from the moving camera of a self-driving car is a challenging problem. In this work, we present a system that improves the localization of static objects by jointly-optimizing the components of the system via learning. Our system is comprised of networks that perform: 1) 5DoF object pose estimation from a single image, 2) association of objects between pairs of frames, and 3) multi-object tracking to produce the final geo-localization of the static objects within the scene. We evaluate our approach using a publicly-available data set, focusing on traffic lights due to data availability. For each component, we compare against contemporary alternatives and show significantly-improved performance. We also show that the end-to-end system performance is further improved via joint-training of the constituent models. Code is available at: \url{https://github.com/MedChaabane/Static_Objects_Geolocalization} . 

\end{abstract}

\section{Introduction}

Many self-driving vehicle systems rely on a high-definition (HD) map to ensure safety, driving comfort and legal conformance. Unlike a standard navigation map, an HD map contains detailed 3D structure such as LiDAR point clouds, as well as the precise position and semantics of traffic signs, lights, lanes, and other road markings. One challenge when using an HD map is some portion or set of objects in the map may be out-of-date with changes that occur in the world. The safety of self-driving systems is improved when on-board perception systems not only detect and track dynamic actors in the scene, but also perceive the static traffic-control objects. This allows the system to combine the benefits provided by both perception and mapping for traffic-control features -- timeliness of real-time perception, human-verified accuracy of the map.

In this work, we present a method for 3D detection, tracking, and localizing spatially-compact static objects (such as signs and traffic lights) from a single camera of a self-driving car. We assume that each frame of video can be associated with a reasonable ego-pose of the camera, as is readily available in open-source self-driving data sets. Our method consists of neural networks that address each of the main components of the system, combined to allow joint-optimization via learning to improve overall performance. Given the problem domain, we constrain the solution space to online methods.

The top-level model takes a pair of geo-located video frames as input and outputs a set of localized objects (5 Degree-of-Freedom, or ``5D" poses). For each input image, a sub-network performs 5D pose regression for each detected object. Detected objects are represented with both appearance and pose information for learning how to associate them between frames. We employ an existing object detector, but propose new networks for single-image object pose regression and cross-image object matching. The system applies these networks in a multi-object tracking paradigm to produce robust 5D locations for the set of tracked objects in a video sequence.

We evaluate the performance of the proposed approach on traffic lights due to availability of data. In principle, this method could be applied to other static object types as well. In summary, our main contributions are: (i) a novel pose regression network for estimating 5D poses of static objects from geolocated RGB inputs, shown to outperform contemporary methods, (ii) a novel method for matching objects between pairs of video frames combining multi-resolution appearance features and geometric features from our pose regression network, (iii) the formulation of multi-object tracking of static objects using these models, and (iv) an evaluation comparing the performance of the individual components against contemporary alternatives, and also showing the benefit to the system-level performance of jointly-optimizing the models with a multi-task loss function.


\section{Related Work}

Localizing street-level objects using multi-view geometry has been the focus of important prior work. Hebbalaguppe et al. \cite{hebbalaguppe2017telecom} proposed an automatic system to update telecom inventory using stereo-vision distance estimation with a SIFT feature matching algorithm, applied to Google street view images. Krylov et al. \cite{krylov2018automatic} combined monocular depth estimation and triangulation to enable automatic localization of static objects. The same authors extended their approach by adding LiDAR data for object segmentation, triangulation, and monocular depth estimation for traffic lights \cite{krylov2018object}. Zhang et al. \cite{zhang2018using} proposed a method for mapping roadside utility poles from street view images using a CNN-based object detector followed by a line-of-bearing method for object-localization.

In contrast to these works, we hypothesize that an end-to-end trainable system will perform better when compared to systems using disjoint components \cite{ruder2017overview}. Prior works commonly use deep learning to detect objects in imagery, but then employ distinct secondary processes to track or otherwise associate observations across images, lacking the full support of information from the object detection model. Consequently, as the number of nearby objects increases, geometric-only techniques can fail because of the inherent spatial uncertainty of the features. In our approach, we make the assumption that strong similarities can be derived from complementary visual and geometric features, and that jointly learning these features in a single end-to-end system has additional performance benefits.

The prior work most closely related to ours is by Nassar et al. \cite{nassar2019simultaneous}, who proposed an end-to-end trainable object geo-localization architecture. A pair of images is fed to their architecture: objects are first detected in the image pairs, then matching projections are learned, and finally the geo-coordinates of the objects are predicted. Our work shares a commitment to an end-to-end approach, but differs significantly in implementation details. Also, our additional multi-object tracking stage is novel and improves overall performance.

Tracking static objects from a moving camera can be considered a special case of the typical application of tracking moving objects (from either a static or moving camera). Recent research of multi-object tracking primarily follows the tracking-by-detection paradigm. Several different RGB-based approaches belong to this category. One category relies on exploiting re-identification modules \cite{bergmann2019tracking,wojke2017simple,zhang2019robust,sun2019deep,zhu2018online} to accurately match objects between frames. 
Another category uses motion and continuity cues \cite{choi2015nearr,karunasekera2019multiple,milan2017online,wang2019exploit}. Other approaches rely on the 3D properties as well such as shape and approximate depth \cite{scheidegger2018mono,sharma2018beyond}. However, when considering static objects, object poses can be exploited for tracking in a stronger fashion that can be done when tracking dynamic objects. Our method incorporates the features from jointly-learned pose and appearance features to track static objects across video frames.



\section{Proposed Approach}
Our object localization method consists of two models. The first is a pose regression network (\S~\ref{pose-regression-section}) used to estimate the 5D pose of objects present in an RGB image. The second is an object matching network (\S~\ref{section_matchimg}) used to associate objects across a sequence of frames.

Our approach is an online method, so it uses information derived only from past frames, making it suitable for use in self-driving vehicles and other streaming applications. At each given frame $t$, the network produces a set of 2D object detections in the image. For each detection, the 5D pose is estimated. The current-frame detections are associated with tracks of previously-detected objects using the object matching network. For each tracked object, we aggregate the estimated 5D poses over time to compute the final location and rotation. Object locations are aggregated by applying an LSTM network. In this section, we provide details on the two main components, the pose regression network and the object matching network.

\subsection{Pose Regression Network} \label{pose-regression-section}
\begin{figure}[t]
\centering
\includegraphics[height=7.5cm]{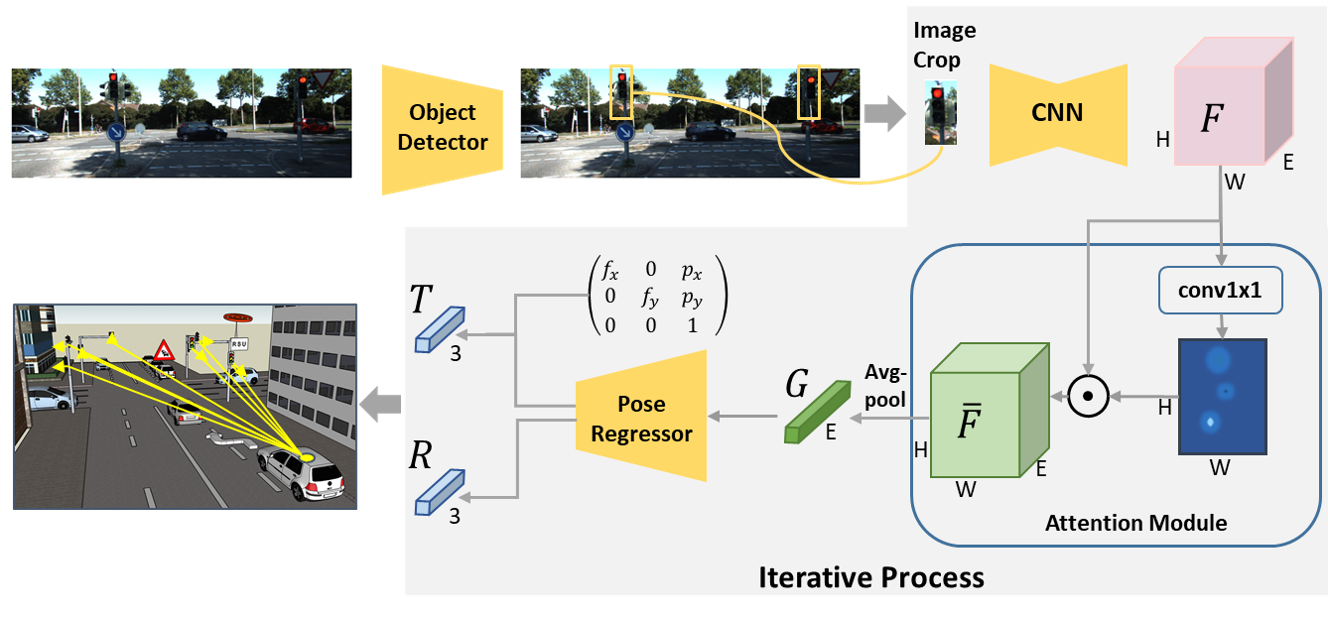}
\caption{\textbf{Single-image Object Pose Regression} Our model first computes bounding boxes (crops) of objects of interest from geolocated images. Each image crop is then processed with an encoder-decoder CNN to generate a feature map, $F$, which is processed by an attention module to yield $\bar{F}$. Using average pooling, we create a fixed-size geometry embedding $G$, which is fed to the pose regressor to output the 5D pose.}
\label{fig:DeepPRR}
\end{figure}

\figurename{~\ref{fig:DeepPRR}} illustrates the architecture of our object pose regression model. Our approach is designed for online processing of a stream of geolocated images, such as those that might be produced by self-driving vehicles. The method is for application to spatially compact static objects, such as traffic lights or signs. We use the term ``spatially compact" to distinguish such objects from things like lane lines or road edge boundaries. As static objects of interest are tracked across frames, the per-frame pose estimates are used not only to refine the final 5D pose of the object, but also to help disambiguate matching objects across frames (see \S~\ref{Feature-sub-network}). Our network outputs 5D object pose vectors $p=[T,R]$ where $T = (T_x, T_y, T_z)$ represents the 3D translation vector of the center of the object in the camera coordinate system and $R= (R_x,R_y)$ represents the unit vector orthogonal to the object (the direction in which traffic light or sign is facing) with respect to the camera coordinate frame. To estimate the pose, we train our network using the euclidean loss $L_{\textrm{trans}}(T,\hat{T})= \|T-\hat{T} \|_2$ for the translation regression, and the log hyperbolic cosine loss  $L_{\textrm{rot}}(R, \hat{R})= \sum_{a \in \{x,y\}} \log(\cosh(R_a - \hat R_a))$ for the rotation regression, where $p=[T,R]$ is the ground truth pose and $\hat p=[ \hat T, \hat R]$ is the estimated pose. Instead of regressing the full translation vector $T$, our pose regression network is trained to regress the $T_z$ component and the object's center position $c = (c_x, c_y)$ in image pixel space. This formulation provides better invariance to camera parameters. We use projective geometry to recover the full translation vector $T_a = (c_a-p_a)T_z / f_a$ for $a \in \{x,y\}$, where $f_x$, $f_y$ are the camera focal lengths, and ($p_x$, $p_y$) is the camera principal point offset.

Our pose regression network is a two-staged network. The first stage is a typical 2D object detection network \cite{bai2017deep,redmon2016you,xiong2019upsnet}. We pad the bounding boxes of the detected objects by $N_p$ pixels for each side to include more context and to take into account slight errors coming from the object detector model. Features from within each padded bounding box (``image crop") are used in the second stage to estimate object pose.

\subsubsection{Geometry Embedding}
The image crop is fed into an encoder-decoder network that maps an image of size $H \times W \times 3$ into a feature map $ F \in \mathbb{R}^{H \times W \times E} $. Each pixel of the feature map is an $E$-dimensional vector representing the appearance information of the input image crop at each pixel location. From the feature map $F$, we derive the embedding of the image crop as follows. We employ a spatial attention mechanism to focus the embedding on the most salient parts of the image crop. The spatial attention distribution $ a \in \mathbb{R}^{H \times W } $ is learned using $1\times1$ convolutions from the extracted feature maps $F$. The spatial attention map $a$ is then normalized using softmax of the responses:
\begin{equation} \label{eqSpatialAttention}
\Bar{a} = \frac{\textrm{exp}(a)}{\sum^H_{i=1} \sum^W_{j=1} \textrm{exp}(a_{i,j})}
\end{equation}
The normalized spatial attention map $\Bar{a}$ is applied to weight the feature map $F$ to generate the attention-weighted feature map $\Bar{F} = \textrm{rep}(\Bar{a}) \odot F$ (we replicate $\Bar{a}$ for $E$ times to match the size of $F$). Average pooling is then applied to $\Bar{F}$ to obtain the geometry embedding $G \in \mathbb{R}^{E} $.

\subsubsection{Pose Regressor} 
The pose regressor transforms the geometry embedding $G$ into 5D pose estimates for each object crop in the input image. The pose regressor is composed of a rotation and a translation branch, each composed of fully connected layers. The rotation branch estimates the rotation vector $R$ and is normalized before computing the loss. The translation branch estimates the $T_z$ component of the translation vector and the object's center position $c = (c_x, c_y)$. The network is trained by minimizing the loss $L_{\textrm{pose}} = L_{\textrm{rot}}  + \beta L_{\textrm{trans}}$.

\subsection{Object Matching Network } \label{section_matchimg}
\begin{figure}[t]
\centering
\includegraphics[height=7cm]{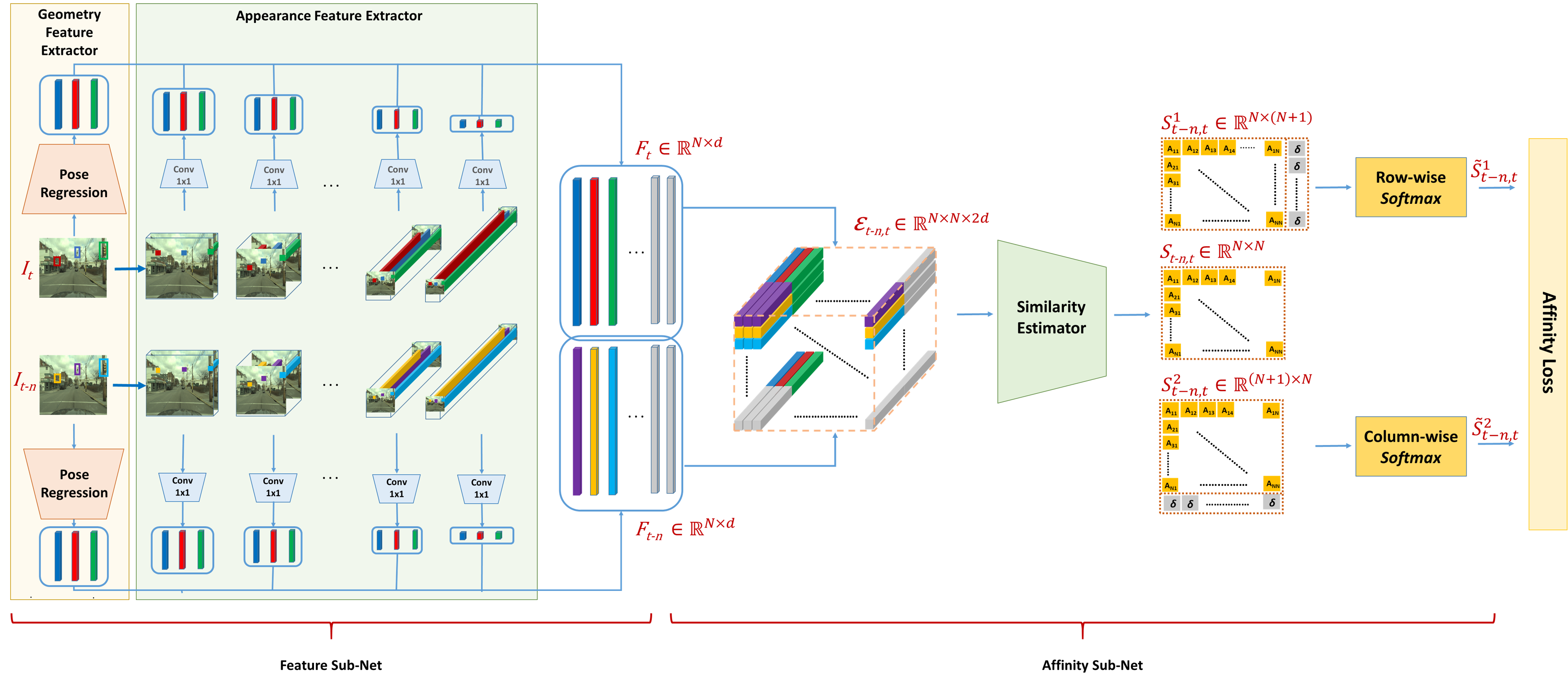}
\caption{\textbf{Object Matching Network.} A pair of images $n$ frames apart, $I_t$ and $I_{t-n}$, along with the detected 2D bounding boxes, are input to the network. The feature sub-network extracts a $d$-dimensional vector encoding pose and appearance information for each detected object in each frame. The affinity sub-network uses these to produce affinity estimations, matching objects across the two frames.}
\label{fig:DeepOM}
\end{figure}

The object matching network is responsible for associating objects between pairs of frames, allowing the system to track objects through the video sequence. At a high level, our architecture follows the Deep Affinity Network (DAN) of \cite{sun2019deep}, but we make modifications to add pose features into the embeddings. With these changes, the object matching network supports joint learning of pose, appearance, and similarity to improve static object tracking. A summary of the architecture is presented in \figurename{~\ref{fig:DeepOM}}.

\subsubsection{Data Preparation and Encoding}
A pair of images $n$ frames apart, $I_t$ and $I_{t-n}$, are input to the object matching network along with the sets of bounding boxes of the detected objects, $B_t=[b^t_1,b^t_2,...,b^t_{N1}]$ and $B_{t-n}=[b^{t-n}_1,b^{t-n}_2,...,b^{t-n}_{N2}]$ respectively, with $ 1 \leq N_1,N_2 \leq N$ where N is the maximum number of allowed detected objects in any frame. In order to provide more robustness during inference, the matching network is trained using image pairs separated by a variable amount of time. The lower bound of the interval is a single frame of separation. The upper bound is a number of frames representing a few seconds of time, to allow capturing the situation where the camera has moved significantly between two observations of the same object. When generating the training data, we sample uniformly from the range between the lower and upper time intervals (in number of frames between image pairs), and the training data is expressed as $X_{train} = \{ (I_t, I_{t-n})  \mid n  \in [1, n_{max}] \}$, where $n_{max}$ is the maximum frames of separation between image pairs. Each image in a pair is resized to a fixed-size. For each training pair, we create the ground truth matching matrix $M_{t-n,t} \in \{0, 1\}^{(N+1) \times (N+1)} $ which contains matching scores between $N_1$ objects of frame $I_{t-n}$ (in the rows) and $N_2$ objects of frame $I_t$ (in the columns). We add $N-N_1$ nonexistent objects to $I_{t-n}$ and $N-N_2$ nonexistent objects to $I_{t}$ in order to obtain a fixed-size matching matrix. These additional rows and columns are filled with zeros. An element $M_{t-n, t}[i, j]$ from the matching matrix encodes the association between the object observations $b^{t-n}_i$ and $b^{t}_j$. A value of $1$ encodes an association, meaning that the observations pertain to the same physical object. Entities entering and leaving the scene are encoded with $M_{t-n, t}[N+1, j]=1$ and $M_{t-n, t}[i, N+1]=1$, respectively.

\subsubsection{Feature Sub-network} 
\label{Feature-sub-network}
The feature sub-network extracts the compact features used to associate objects between image pairs. The pair of frames $(I_{t-n}, I_t)$ are fed in parallel to the feature sub-network where the two branches share the same set of weights. This sub-network is composed of a geometry feature extractor (yellow box in \figurename{~\ref{fig:DeepOM}}) and an appearance feature extractor (green box in \figurename{~\ref{fig:DeepOM}}). The underlying idea of our feature sub-network is that we can compute the affinity scores between objects based on visual and geometric cues.


We are mainly focusing on autonomous driving scenes, where the video frames are from a monocular camera mounted on a car moving on the road plane, and the tracked targets are static objects near the road. Thus, geometry features that describe the location and rotation of objects can be helpful to discriminate between objects. Benefiting from reliable pose estimation, we expect that the same physical object in the 2 frames $I_{t-n}$ and $I_t$  will have similar estimations of location and rotation in a \textit{common reference frame}. 

Thus, from any frame $I_t$, we use our pose regression network to output the estimated location and rotation of the detected object. The estimated pose is then transformed into the camera coordinates system of a common reference frame $I_{ref}$; in our implementation we chose the reference frame to be the first frame for each video.



The geometry embedding $G$ used in the pose regression network contains information about the geometry of the objects as well. Thus, we concatenate the features of $G$ with the 6 pose values described above to construct $f_{g,i} \in \mathbb{R}^{6+E} $ geometry feature descriptor for the $i^{\textrm{th}}$ detected object. 

\begin{table}[b]
\begin{center}
\caption{Details on the architecture of the appearance sub-network used in the object matching network. The layers used in the final embedding are denoted in the column ``Label'' as $f_{R_n} | n \in[1,10]$.}
\label{table:arch-appearance}
\begin{tabular}{|ccccccc|}
\hline
\multicolumn{2}{|c}{Layer} & Output size & Kernel size & Stride  &  Receptive field & Label\\
\hline \hline

1& $3 \times $ conv & $H \times W \times 3$ & $ 3 \times 3$ & 1   & 7 & -\\
 &Max Pool & $ \sfrac{H}{2} \times \sfrac{W}{2} \times 64$ & $ 3 \times 3$ & 2  &9&-\\
4&$2 \times $ conv & $\sfrac{H}{2} \times \sfrac{W}{2} \times 64$ & $ 3 \times 3$ & 1  &17&-\\
&Max Pool& $\sfrac{H}{4} \times \sfrac{W}{4} \times 128$ & $ 3 \times 3$ & 2  &21&$f_{R_1}$\\
6&$3 \times $ conv & $\sfrac{H}{4} \times \sfrac{W}{4} \times 128$ & $ 3 \times 3$ & 1  &45&-\\
&Max Pool& $\sfrac{H}{8} \times \sfrac{W}{8} \times 256$ & $ 3 \times 3$ & 2  & 53&$f_{R_2}$\\
9&$2 \times $ conv & $\sfrac{H}{8} \times \sfrac{W}{8} \times 256$ & $ 3 \times 3$ & 1  &85&$f_{R_3}$\\
11&$1 \times $ conv & $\sfrac{H}{8} \times \sfrac{W}{8} \times 512$ & $ 3 \times 3$ & 1  &101&-\\
&Max Pool& $\sfrac{H}{16} \times \sfrac{W}{16} \times 512$ & $ 3 \times 3$ & 2  &117&$f_{R_4}$\\
12&$3 \times $ conv & $\sfrac{H}{16} \times \sfrac{W}{16} \times 512$ & $ 3 \times 3$ & 1 &213&$f_{R_5}$\\
15&$2 \times $ conv & $\sfrac{H}{16} \times \sfrac{W}{16} \times 512$ & $ 3 \times 3$ & 1  &277&$f_{R_6}$\\
17&$2 \times $ conv & $\sfrac{H}{16} \times \sfrac{W}{16} \times 512$ & $ 3 \times 3$ & 1  &341&-\\
&Max Pool& $\sfrac{H}{32} \times \sfrac{W}{32} \times 512$ & $ 3 \times 3$ & 2  &373&$f_{R_7}$\\
19&$3 \times $ conv & $\sfrac{H}{32} \times \sfrac{W}{32} \times 512$ & $ 3 \times 3$ & 1  &565&$f_{R_8}$\\
22&$3 \times $ conv & $\sfrac{H}{32} \times \sfrac{W}{32} \times 1024$ & $ 3 \times 3$ & 1  & 757&-\\
&Max Pool& $\sfrac{H}{64} \times \sfrac{W}{64} \times 1024$ & $ 3 \times 3$ & 2  & 821&$f_{R_9}$\\
25&$2 \times $ conv & $\sfrac{H}{64} \times \sfrac{W}{64} \times 1024$ & $ 3 \times 3$ & 1  &1077&-\\
&Max Pool& $\sfrac{H}{112} \times \sfrac{W}{112} \times 1024$ & $ 3 \times 3$ & 2  & 1205&$f_{R_{10}}$\\

\hline
\end{tabular}
\end{center}
\end{table}
Given a monocular imaging system, the objects and close surroundings are expected to maintain their visual appearance over short time spans. To extract appearance features, we employ a convnet inspired by the increased performance of CNNs with smaller filter size ($3 \times 3$) and deeper architectures such as VGG \cite{Simonyan15}. It consists of 26 convolutional layers and 7 max-pooling layers. Each convolutional layer is followed by batch normalization \cite{ioffe2015batch} and a ReLu activation function; see Table~\ref{table:arch-appearance} for more details. 
For each detected object, we extract feature vectors from the object's center location as regressed from our pose regression network. If the center of the $i^{\textrm{th}}$ object is at position $(X_{i,r},X_{i,c})$ in the input frame of size $W \times H$, then for a feature map $\mathcal{F}_j$ of size $W_j \times H_j \times C$, we extract $C$-dimensional feature vector at position $(\frac{X_{i,r}}{H}H_j,\frac{X_{i,c}}{W}W_j )$ as the corresponding feature vector for the $i^{\textrm{th}}$ object. 
This multi-resolution architecture helps to simultaneously capture fine geometric details as well as higher-level semantics of the surroundings. We show that using a multi-resolution feature vector outperforms those using a single receptive field (see \S~\ref{object-matching}). 


After extracting appearance and geometry features for each detected object, we concatenate both to obtain $f_{i}= f_{g,i} \bigoplus f_{a,i} \in \mathbb{R}^{d} $ ($d= A+6+E$) which is a fused feature descriptor for the $i^{\textrm{th}}$ detected object. For each frame, $I_t$, we construct matrix $F_{t} \in \mathbb{R}^{N \times d}$, by padding by rows (filled with zeros) for nonexistent objects to construct fixed-size feature matrices.

\subsubsection{Affinity Sub-network}
Using the extracted feature matrices $F_{t-n} $ and $F_{t} $, we build the tensor $ \mathcal{E}_{t-n,t} \in \mathbb{R}^{N \times N \times 2d} $ where $\mathcal{E}_{i,j,:} = f_{i,t-n} \bigoplus f_{j,t}$ is the concatenation of the feature vectors of the $i^{\textrm{th}}$ object of $I_{t-n}$ and the $j^{\textrm{th}}$ object of $I_{t}$. The tensor  $ \mathcal{E}_{t-n,t} $ contains all possible $N \times N$  concatenations of feature vectors of objects between the two frames. 
This formulation allows us to compute object affinities in a single forward pass. $ \mathcal{E}_{t-n,t} $ is fed to a similarity estimator network composed of 6 layers of $ 1 \times 1$ convolutions. The output of the similarity estimator network is similarity matrix $S_{t-n,t} \in [0,1]^{N \times N}$ where each element $S_{i,j}$ represents the affinity between bounding box $b^{t-n}_i$ and $b^{t}_j$. Note that we use $1 \times 1$ convolutions so that the computation of $S_{i,j}$ is computed using only the feature vectors $f_{i,t-n}$ and $ f_{j,t}$ and will not be affected by other feature vectors. 
To consider objects entering and leaving between the two frames, we construct two matrices $S^1_{t-n,t} \in \mathbb{R}^{N \times (N+1) }$ and $S^2_{t-n,t} \in \mathbb{R}^{(N+1) \times N}$ where we append a column and a row, respectively. These additional rows and columns are filled 
with a basis value $\delta$. Then, we apply column-wise and row-wise softmax to $S^1_{t-n,t}$ and $S^2_{t-n,t}$ respectively to obtain $\tilde{S}^1_{t-n,t}$ and $\tilde{S}^2_{t-n,t}$ which are fed to the affinity loss layer.

\subsubsection{Joint Loss Function}
To train the object matching network, we use the loss function $L_{\textrm{Aff}}$ as the average of losses $L_{1}$ and $L_{2}$ where $L_{1}$ is the error of matching objects detected in $I_{t-n}$ to the objects in $I_{t}$ and $L_{2}$ is the error of matching objects detected in $I_{t}$ to the objects in $I_{t-n}$. The expression of the losses are given by: 
\begin{eqnarray} \label{eqAffinityLoss}
L_{k \in [1,2]} & = & -\frac{1}{N_k} \sum^{N_k}_{i=1} \sum^{N+1}_{j=1}  m_{i,j} \textrm{log}(\tilde{s}^k_{i,j}), \\
L_{\textrm{Aff}} & = & \frac{L_{1} + L_{2}}{2},  
\end{eqnarray}


where $m_{i,j}$, $\tilde{s}^1_{i,j}$ and $\tilde{s}^2_{i,j}$ are the elements in the $i^{\textrm{th}}$ row and $j^{\textrm{th}}$ column of matrices $M_{t-n,t}$, $\tilde{S}^1_{t-n,t}$ and $\tilde{S}^2_{t-n,t}$ respectively. In inference, the similarity score between $i^{\textrm{th}}$ object of $I_{t-n}$ and the $j^{\textrm{th}}$ object of $I_{t}$ is given as the average of $\tilde{s}^1_{i,j}$ and $\tilde{s}^2_{i,j}$.

Training optimizes the joint affinity and pose estimation losses as defined in Eq. \eqref{eqjoint}. The loss of the pose estimation task is computed as the average of the pose losses of all object detected in both frames. Pose and affinity losses are traded-off with a scalar $\lambda$.
 
 \begin{equation} \label{eqjoint}
L_{\textrm{joint}}=   L_{\textrm{Aff}} + \lambda (\frac{1}{N_1+N_2} \sum^{N_1+N_2}_{i=1} L^i_{\textrm{pose}}) 
\end{equation}
 
\subsubsection{Multi-Object Tracking}
Our Multi-Object Tracking (MOT) approach follows the tracking-by-detection paradigm. Given a new frame with the bounding boxes of the detected objects, the tracker computes the similarity scores between the already tracked $m$ targets (each target consists of multiple instances from different frames) and the $n$ newly detected objects using the object matching network. The score matrix is defined as $S = [s_i^j \mid 1 \leq i \leq m \textrm{ and } 1 \leq j \leq n+m]$,  where $s^j_i$ represents the similarity between the $i^{\textrm{th}}$ target and $j^{\textrm{th}}$ detection and it is computed as the maximum over the similarity between the instances of the $i^{\textrm{th}}$ target before frame $t-1$ and the $j^{\textrm{th}}$ detection at current frame $t$, $s^{i+n}_i$ for  $1\leq i \leq m $ represents the likelihood of $i^{\textrm{th}}$ target to not being matched to any of the new detected objects at frame $t$ and is computed as the average of the values at last column in $\tilde{S}^1_{t-n,t}$ for the instances of $i^{\textrm{th}}$ target and $s^j_i = -\infty$ for $j > n$ and $j \ne i$. Finally, the widely-used Hungarian algorithm \cite{kuhn1955hungarian} is adopted to derive the optimal assignments. 


\section{Experiments}
\subsection{Datasets}
We constructed the Traffic Lights Geo-localization (TLG) data set. TLG is derived from nuScenes \cite{caesar2019nuscenes}, a popular open-source data set for autonomous driving. The nuScenes data contains 1000 scenes of 20 seconds (at 12Hz video rate), filmed in two cities (Boston and Singapore), in both night and day, and with three weather conditions (rain, sun and clouds). Each scene comes with data from six cameras placed at different angles on the car. 

We selected those scenes within road intersections containing traffic lights (TLs). For each scene in the nuscenes data set, and for each video clip from one of the six cameras, we iterated through key frames (2Hz), selecting TLs within 100 meters of the camera location. Each TL location was transformed from world coordinates to camera coordinates, and then into 2D homogeneous image coordinates, using the provided extrinsic and intrinsic camera calibration parameters. We filter TL locations not visible to the camera. Finally, scenes are selected only if, at least one TL is visible in 5 different key frames in one of the six cameras. With this process, we ended up with 348 scenes for training and 56 scenes for testing. On average, two traffic lights appear per image.

In the TLG data, each video clip (from different cameras) in each scene contains 240 RGB images (including 40 key frames) with resolution of $1600 \times 900$. Images are augmented with camera pose information and camera metadata, including information about each visible TL: unique ID, 5D pose in world coordinates, 5D pose in the camera coordinates of the first frame, and TL type (horizontal or vertical). 

We created three sub-datasets for our main tasks, one each for pose, matching, and tracking. The ``Traffic Lights 5D Pose" data contains around 66,000 snippets of TLs (60,000 for training and 6,000 for testing) along with their 5D poses. The ``Traffic Lights Matching" data contains 200,000 pairs of images (170,000 for training and 30,000 for testing) along with bounding boxes of TLs and ground truth matching matrices between the two images. Average elapsed time between image pairs in the Traffic Lights Matching data set is 1.4 seconds (the maximum frames of separation between image pairs $n_{max}$ is set to 35) and on average, four traffic lights appear per image. The ``Multi-Traffic Lights Tracking" (MTLT) data provides a detection and annotation file for each video following the format of \cite{milan2016mot16}.

We evaluated several other potential sources of data that we hoped could be used to evaluate our static object localization approach. Unfortunately, beyond nuScenes, we were unable to find other useful data sets.

\subsection{Implementation details}
We implement our proposed approach using PyTorch \cite{paszke2019pytorch}. All experiments were run on an Ubuntu server with an Nvidia TitanX GPU with 12GB of memory. The performance comparison of contemporary methods for all tasks evaluated in this work were produced using the original authors' publicly-available code.

In the pose regression network, our 2D object detector is the same as used in PoseCNN \cite{xiang2017posecnn}. It is pre-trained on COCO \cite{lin2014microsoft} and Mapillary \cite{neuhold2017mapillary} datasets. The bounding box padding, $N_p$, is set to be between 5-25 pixels, scaled based on the bounding box. The architecture used to extract feature map $F$ is composed of a Resnet-18 encoder followed by 4 up-sampling layers as decoder. The geometry embedding dimension $E$ is set to 128. The weight factor $\beta$ is set to 0.1. Our pose regression network is trained using SGD for 40 epochs with a momentum of 0.9, and a weight decay of 0.0005.

For the object matching network, the maximum number of tracked objects, $N$, is set to 30 and $\delta$ is set to 8. The frames were resized to 896 × 896. By experimental evaluation, the optimal dimensions of the appearance features vectors $f_{R1},f_{R2},\dots,f_{R10}$ are set to 100, 80, 70, 60, 50, 40, 30, 30, 20 and 20 respectively, which results in a 634-dimensional (500 + 6 + 128) feature descriptor for each detected object. The object matching and pose regression networks are jointly trained for 130 epochs with a momentum of 0.9, a weight decay of 0.0008, and $\lambda$ is 0.005. The pose network is initialized to pre-trained weights. 

\subsection{5D Pose Estimation} 
\label{5D-pose-section}
Many state-of-the-art methods for object pose estimation \cite{li2018deepim,oberweger2018making,peng2019pvnet,zakharov2019dpod} use 3D models of the objects. These methods do not work well for our application because of the presence of multiple types and sizes of TLs (and other static objects of interest) in real-world scenarios. Thus, we compared our model to those which take RGB images as input and regress directly 5D poses such as PoseNet \cite{kendall2015posenet} and PoseCNN \cite{xiang2017posecnn}. To make the comparison fair, all methods use the same object detector \cite{long2015fully} as in PoseCNN, and we fine-tune both PoseNet and PoseCNN on our training data with the same loss function used to train our pose regression network. Table~\ref{table:pose1} presents a comparison of our pose regression model against PoseNet and PoseCNN on the Traffic Lights 5D Pose data.

Our single-image pose regression network outperforms both PoseNet and PoseCNN. As expected, TLs far away from the camera can be challenging to locate accurately. All methods have considerably lower pose errors when evaluating only on TLs within 20 meters. In the full data set, TLs can be up to 100 meters away from the camera. We show in a later discussion of end-to-end performance (see Table~\ref{table:analysis-axes}) that most of the translation error is concentrated in the depth axis, $T_z$.

To understand the effects of the attention module and joint training strategy, we compared the performance of three variants of our pose regression network as shown in Table~\ref{table:pose1}. The inclusion of the attention module reduces the rotation and translation errors. This shows how focusing on some regions in the image crop helps our model to extract a better representation for 5D pose regression. We also see that training the pose regression and object matching networks jointly improves pose regression performance.

\setlength{\tabcolsep}{4pt}
\begin{table}
\begin{center}
\caption{Pose regression ablation study. 
In ``w/o Attention" we removed the attention module of the pose regression ($\bar F =F$). In ``Joint Training", the regression model is trained jointly with the object matching model to minimize loss function $L_{\textrm{joint}}$ in Eq. \eqref{eqjoint}. ``Baseline" indicates training the model as described, as a stand-alone network}
\label{table:pose1}
\begin{tabular}{|c|cc|cc|c|}
\hline 
Model & \multicolumn{4}{c|}{ 5D Pose Errors (mean/median)} & Run time\\
\hline \hline
&  \multicolumn{2}{c|}{ All objects} & \multicolumn{2}{c|}{ Near ($\leq 20 \textrm{m}$) objects} & sec/frame \\
& Translation (m) & Rotation ($^{\circ}$) & Translation (m) & Rotation ($^{\circ}$) & \\
Ours (w/o Attention) & 4.95 / 3.93 & 17.68 / 10.51 & 3.02 / 2.24 & 16.26 / 7.64 & 0.05 \\
Ours (Baseline) & 4.67 / 3.61 & 17.00 / 9.70 & 2.64 / 1.83 & 14.74 / 6.24 & 0.05\\
Ours (Joint Training) & \textbf{4.43} / \textbf{3.39} & \textbf{15.97} / \textbf{9.16} & \textbf{2.51} / \textbf{1.70}& \textbf{14.21} / \textbf{6.08} & 0.05\\
PoseNet \cite{kendall2015posenet} & 7.25 / 5.83 & 28.47 / 21.82 & 5.36 / 4.48 & 24.31 / 18.23 & \textbf{0.04} \\
PoseCNN \cite{xiang2017posecnn} & 5.54 / 4.47 & 19.63 / 11.35 & 3.68 / 2.91 & 18.04 / 8.86 & 0.11\\
\hline
\end{tabular}
\end{center}
\end{table}
\setlength{\tabcolsep}{1.4pt}

\subsection{Object Matching} \label{object-matching}

\begin{figure}[b]
\begin{floatrow}

\capbtabbox{%

  \begin{tabular}{|c|cc|}
\hline
Object Matching Feature Extractor & mAP &  Runtime\\
\hline
\hline
Resnet-50 \cite{he2016deep}& 0.744 & 0.1 \\
VGG-16 \cite{Simonyan15} & 0.824 & \textbf{0.08} \\
AFE (RFs $\leq$ 213 only) & 0.857 & 0.11 \\
AFE (RFs $>$ 213 only) & 0.839 & 0.12 \\
AFE & 0.873 & 0.12 \\
GFE (5D only) & 0.825 & \textbf{0.08} \\
GFE (5D + $G$) & 0.831 & \textbf{0.08} \\
AFE + GFE & 0.912 & 0.14 \\
AFE + GFE (Joint Training) & \textbf{0.928} & 0.14\\
\hline
\end{tabular}
}{%
  \caption{Object matching network ablation study. AFE uses only the appearance features. GFE uses only the geometry features. For AFE, also shown is the impact on receptive field (RF) sizes. For GFE, we show with and without including the pose regression feature vector $G$
}%
\label{table:feature-extractor}
}
\ffigbox{%
  \includegraphics[height=4.0cm]{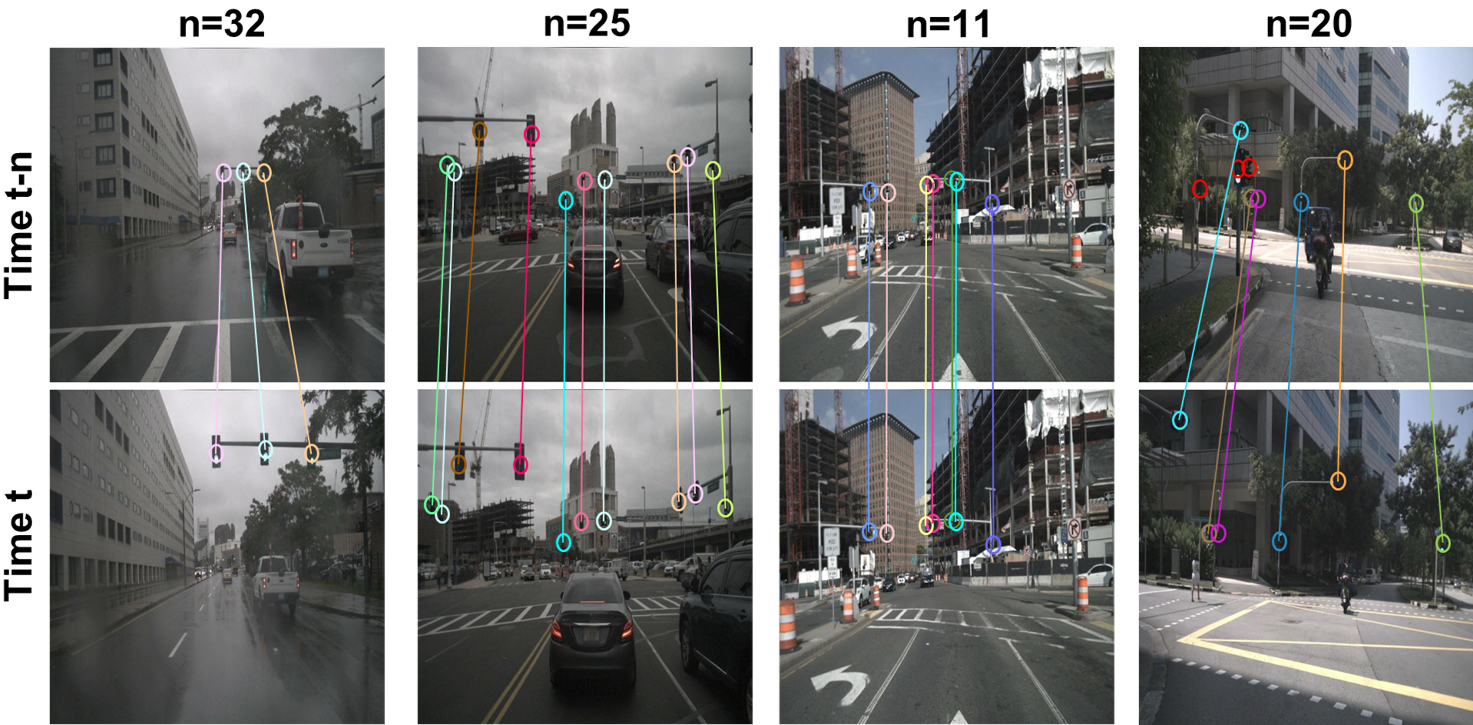}
}{%
  \caption{Object matching examples. Each column of the figure shows a pair of frames separated by n frames. Object matching remains robust to illumination and weather conditions and existence of multiple similar TLs in the frames.}%
  \label{fig:Data-association}
}
\end{floatrow}
\end{figure}

To highlight the impact of the feature sub-network of the object matching network, we report matching accuracy after changing the feature extractor component in Table~\ref{table:feature-extractor}. In this ablation study, we measure the impact of using only appearance features, only geometric features, using both appearance and geometric features, and joint training. Additionally, we measure variants of the appearance features when larger or smaller receptive fields are used, and we show variants of the geometric features when using only the 5D values or when combining the 5D values with the vector $G$ from the pose regression network. 

In Table~\ref{table:feature-extractor} and in the following text, ``AFE" will indicate using only appearance features and ``GFE" will indicate using only geometric features in the object matching network. AFE outperformed single RF based architectures (Resnet-50 and VGG-16) by more than 4.9 percentage points, which demonstrates the benefit of multi-resolution networks for our application. We found that appearance features extracted from small RFs perform better than those extracted from larger RFs, as illustrated when comparing AFE (RFs $ > 213$) and AFE (RFs $\leq 213$). This fact is supported by comparing Resnet-50 (RF size = 483) and VGG-16 (RF size = 212), where VGG outperforms Resnet-50. Combining features from both small and large RFs (AFE) results in mAP gain of 1.6 percentage points. This can be explained by the fact that features from small RFs will focus on low level information such as color, texture, and shape, while features from large RFs will have richer contextual information that can be beneficial in some challenging cases.

By comparing performance of AFE and GFE, we can conclude that appearance is more important than geometry for our object matching network. However, including the geometry cues helps to increase the mAP by 3.9 percentage points over appearance alone. We argue that the advantage gained from geometry features come when TLs look similar and are close in image space. In those cases, TLs will also have similar backgrounds and thus produce similar appearance embeddings. The joint training strategy provides the remaining improvements, increasing the object matching network's mAP by 1.6 percentage points when compared to stand-alone training.

\figurename{~\ref{fig:Data-association}} shows examples of the object matching network's output from our Traffic Lights Matching data. We observe that the association appears robust to illumination and weather conditions. Also, even with the existence of multiple similar looking TLs at very close locations in the image space, the network is able to correctly associate the TLs. The chosen examples in \figurename{~\ref{fig:Data-association}} are random. We noted similar level of performance by the object matching network for all the examples we tested.

\subsection{Multi-Object Tracking}

\setlength{\tabcolsep}{4pt}
\begin{table}
\begin{center}
\caption{Comparison of our method and contemporary MOT trackers on the MTLT test sequences. We utilise the standard MOT metrics \cite{bernardin2008evaluating}: MOTA (multi-object tracking accuracy), MOTP (multi-object tracking precision), MT (number of mostly tracked trajectories), ML (number of mostly lost trajectories), IDS (number of identity switches) and FPS (frame per second). $\uparrow$ and $\downarrow$ indicate higher or lower values are preferred}
\label{table:tracking}
\begin{tabular}{|c|cccccc|}
\hline 
Method & MOTA $\uparrow$ & MOTP $\uparrow$ & MT $\uparrow$& ML $\downarrow$ & IDS $\downarrow$ & FPS $\uparrow$\\
\hline \hline

DMAN \cite{zhu2018online} & 80.79 & 82.40 & 61.12 & 12.91 & 103 & 3.3\\
DeepSORT \cite{wojke2017simple} & 77.69 & 77.81 & 56.34 & \textbf{9.41} & 69 & \textbf{17.2} \\
Tracktor++ \cite{bergmann2019tracking} & 83.31 & \textbf{86.73} & 66.54 & 9.71 & 82& 2.64 \\
Ours (GFE) & 74.12 & 75.32 & 51.17 &20.64 & 162 & 9.7 \\
Ours (AFE) & 81.29 & 82.18&  62.37& 12.66& 96 & 6.1\\
Ours (GFE + AFE) & \textbf{85.52} & 85.14 & \textbf{69.57} & 10.79 & \textbf{61} & 5.3\\

\hline
\end{tabular}
\end{center}
\end{table}
\setlength{\tabcolsep}{1.4pt}
We evaluate the performance of our tracker using MOT metrics and compare its performance with three contemporary online MOT algorithms that are known to have reproducible results with publicly available code (Table~\ref{table:tracking}). By only using appearance features (AFE), our tracker achieves 81.29 in terms of MOTA which is higher than appearance-based trackers (DMAN and DeepSORT), demonstrating the strength of our multi-resolution appearance features. By using only geometry features (GFE), our tracker achieves 74.12 in MOTA. By using both appearance and geometry features, the tracking accuracy is increased to 85.52 in MOTA, out-performing the other methods. Our tracker is twice as fast as Tracktor++, which has somewhat similar performance for many of the metrics other than IDS, where our method is much better.

Our tracker's performance as captured by the MT metric is significantly better, suggesting our tracker generates more integrated trajectories by combining geometry and appearance cues. Similarly, our tracker's identity switches (IDS) value of 61 is best. Both MT and IDS are critical metrics when the output of the tracker is used to generate an aggregated pose estimate, as in our application, as we present in the following section.

\subsection{Object Geo-localization}

The end goal for our application is geo-locating static objects for HD Maps. We evaluate performance in this regard by comparing predicted and ground truth geo-locations of traffic lights in the TLG data set. We compare our proposed approach with MRF-triangulation \cite{krylov2018automatic} and SSD-ReID-Geo \cite{nassar2019simultaneous}. By analyzing the errors of different methods (Table \ref{table:analysis-axes}), we note that errors along Z-axis (depth) are considerably higher than errors along X and Y axes, which is typical for monocular vision-based systems. When localizing traffic lights, errors along Z-axis are less troubling than lateral or vertical errors. This is because the perception of whether or not a traffic light pertains to the self-driving car (i.e., the lane the car is in) is more affected by its horizontal position above the road than the depth along the roadway. A lateral error of 2m could cause confusion about which lane the light controls. On the other hand, a depth error of a few meters is unlikely to cause such confusion. Our method shows a median error in the $X$ and $Y$ axes of less than 20cm, and mean error within 25cm. The median depth error ($Z$ axis) of about 1.5m is well-within the accuracy bounds of the problem domain.
\setlength{\tabcolsep}{4pt}
\begin{table}
\begin{center}
\caption{Translation Error (TE) along X, Y and Z axes
}
\label{table:analysis-axes}
\begin{tabular}{|c|ccc|ccc|ccc|}
\hline 
 Model & \multicolumn{3}{c|}{ TE along X-axis (m)} & \multicolumn{3}{c|}{ TE along Y-axis (m) } & \multicolumn{3}{c|}{ TE along Z-axis (m) }\\
\hline \hline
 & Mean &  Median & Std & Mean & Median & Std & Mean & Median & Std \\
Ours & \textbf{0.25}& \textbf{0.16}& 0.15& \textbf{0.23}& \textbf{0.15}& 0.14& \textbf{2.24}& \textbf{1.47}& 1.28\\
MRF-triangulation & 0.31 & 0.24 & 0.12& 0.35 & 0.27 & 0.15& 4.75 & 3.89& 1.92\\
SSD-ReID-Geo & 0.64& 0.51& 0.37 & 0.51& 0.45&  0.33& 3.77 & 2.85 & 1.68\\
\hline
\end{tabular}
\end{center}
\end{table}

\setlength{\tabcolsep}{1.4pt}

We computed object-based precision/recall using two distance thresholds, 2m Euclidean distance and 3 units of Mahalanobis distance. In this case, 3 units of Mahalanobis distance corresponds to an ellipse defined with semi-axes: x=0.4, y=0.39, and z=3.84 meters. The advantage of the Mahalanobis distance is that it provides much tighter thresholds in the X and Y axes while allowing more tolerance in depth, making it more suitable for our application.

\begin{figure}
\centering
\includegraphics[height=4.6cm]{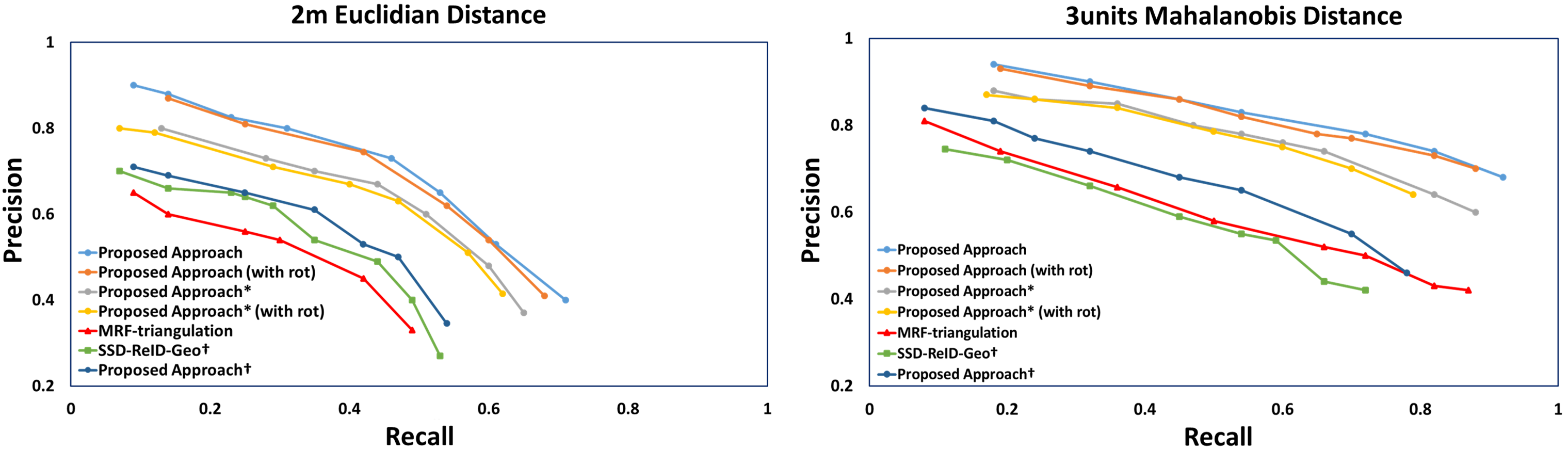}
\caption{Comparison of the performance of our approach for static object geo-localization against MRF-triangulation \cite{krylov2018automatic} and SSD-ReID-Geo \cite{nassar2019simultaneous}. An estimated geo-location is a true positive if it is within a threshold distance of a ground truth point. Methods marked with * use only key frames (2fps) for testing, methods marked with $^\dagger$ are tested with only frame pairs, and ``with rot" means that true positives must also be within $20^{\circ}$ of the true orientation.}
\label{fig:geo}
\end{figure}

\figurename{~\ref{fig:geo}} compares the precision/recall of our approach against MRF-triangulation and SSD-ReID-Geo. Our approach leads to more accurate geolocalizations than the other methods. Our approach outperforms MRF-triangulation thanks to the efficiency of our pose regression model over the depth estimation in \cite{krylov2018automatic}, and the joint learning employed by our approach. SSD-ReID-Geo uses only pairs of frames when estimating object poses. For a fair comparison, we also tested our approach using only frame pairs (\figurename{~\ref{fig:geo}}, denoted with $^\dagger$). Our approach outperforms SSD-ReID-Geo, even with this restriction. We also observe that using the Mahalanobis distance, the PR curve of SSD-ReID-Geo becomes lower than MRF-triangulation due to the added restrictions along X and Y axes.

When adding a rotation error component to the definition of a true positive (i.e., within the distance threshold and within the angular threshold of $20^\circ$), there is only a slight lowering of performance, indicating that our 5D regression performs well for both translation and rotation components.


\section{Conclusions}
This paper proposes an end-to-end method for 5D detection, tracking, and localization of spatially-compact static objects from a single camera of a self-driving car. We showed jointly optimizing the pose regression and object matching models improves 5D pose estimation, tracking and geo-localization simultaneously. Future plans include sharing features between the 2D object detector and the object matching network, which will provide opportunities for further joint optimization and inference speed-up. We also aim to replace the only non-differentiable module of our approach -- the Hungarian algorithm -- with  an equivalent differentiable network to allow complete end-to-end learning. In this work, we were limited to evaluate performance of our approach on traffic lights, application to other static compact object (signs, etc.) requires creating or identifying new data sets.

\bibliographystyle{unsrt}  
\bibliography{references}  

\begin{thebibliography}{10}

\bibitem{hebbalaguppe2017telecom}
Ramya Hebbalaguppe, Gaurav Garg, Ehtesham Hassan, Hiranmay Ghosh, and Ankit
  Verma.
\newblock Telecom {Inventory Management via Object Recognition and
  L}ocalisation on {G}oogle {Street View I}mages.
\newblock In {\em 2017 IEEE Winter Conference on Applications of Computer
  Vision (WACV)}, pages 725--733. IEEE, 2017.

\bibitem{krylov2018automatic}
Vladimir~A Krylov, Eamonn Kenny, and Rozenn Dahyot.
\newblock Automatic {Discovery and Geotagging of Objects from Street View
  I}magery.
\newblock {\em Remote Sensing}, 10(5):661, 2018.

\bibitem{krylov2018object}
Vladimir~A Krylov and Rozenn Dahyot.
\newblock Object {G}eolocation using {MRF} {Based Multi-Sensor F}usion.
\newblock In {\em 2018 25th IEEE International Conference on Image Processing
  (ICIP)}, pages 2745--2749. IEEE, 2018.

\bibitem{zhang2018using}
Weixing Zhang, Chandi Witharana, Weidong Li, Chuanrong Zhang, Xiaojiang Li, and
  Jason Parent.
\newblock Using {Deep Learning to Identify Utility Poles with Crossarms and
  Estimate Their Locations from Google Street View Images}.
\newblock {\em Sensors}, 2018.

\bibitem{ruder2017overview}
Sebastian Ruder.
\newblock An {O}verview of {M}ulti-task {L}earning in {D}eep {N}eural
  {N}etworks.
\newblock {\em arXiv preprint arXiv:1706.05098}, 2017.

\bibitem{nassar2019simultaneous}
Ahmed~Samy Nassar, S{\'e}bastien Lef{\`e}vre, and Jan~Dirk Wegner.
\newblock Simultaneous multi-view instance detection with learned geometric
  soft-constraints.
\newblock In {\em Proceedings of the IEEE International Conference on Computer
  Vision (ICCV)}, 2019.

\bibitem{bergmann2019tracking}
Philipp Bergmann, Tim Meinhardt, and Laura Leal-Taixe.
\newblock Tracking without bells and whistles.
\newblock In {\em Proceedings of the IEEE International Conference on Computer
  Vision}, pages 941--951, 2019.

\bibitem{wojke2017simple}
Nicolai Wojke, Alex Bewley, and Dietrich Paulus.
\newblock Simple {Online and Realtime Tracking with a Deep Association M}etric.
\newblock In {\em 2017 IEEE international Conference on Image Processing
  (ICIP)}, pages 3645--3649. IEEE, 2017.

\bibitem{zhang2019robust}
Wenwei Zhang, Hui Zhou, Shuyang Sun, Zhe Wang, Jianping Shi, and Chen~Change
  Loy.
\newblock Robust multi-modality multi-object tracking.
\newblock In {\em Proceedings of the IEEE International Conference on Computer
  Vision}, pages 2365--2374, 2019.

\bibitem{sun2019deep}
ShiJie Sun, Naveed Akhtar, HuanSheng Song, Ajmal~S Mian, and Mubarak Shah.
\newblock Deep {A}ffinity {N}etwork for {M}ultiple {O}bject {T}racking.
\newblock {\em IEEE Transactions on Pattern Analysis and Machine Intelligence},
  2019.

\bibitem{zhu2018online}
Ji~Zhu, Hua Yang, Nian Liu, Minyoung Kim, Wenjun Zhang, and Ming-Hsuan Yang.
\newblock Online {Multi-Object Tracking with Dual Matching Attention N}etworks.
\newblock In {\em Proceedings of the European Conference on Computer Vision
  (ECCV)}, pages 366--382, 2018.

\bibitem{choi2015nearr}
Wongun Choi.
\newblock Near-{O}nline {M}ulti-{T}arget {T}racking {W}ith {A}ggregated {L}ocal
  {F}low {D}escriptor.
\newblock In {\em Proceedings of the IEEE International Conference on Computer
  Vision}, pages 3029--3037, 2015.

\bibitem{karunasekera2019multiple}
Hasith Karunasekera, Han Wang, and Handuo Zhang.
\newblock Multiple {Object Tracking With Attention to Appearance, Structure,
  Motion and S}ize.
\newblock {\em IEEE Access}, 7:104423--104434, 2019.

\bibitem{milan2017online}
Anton Milan, S~Hamid Rezatofighi, Anthony Dick, Ian Reid, and Konrad Schindler.
\newblock Online {Multi-Target Tracking Using Recurrent Neural Networks}.
\newblock In {\em Thirty-First AAAI Conference on Artificial Intelligence},
  2017.

\bibitem{wang2019exploit}
Gaoang Wang, Yizhou Wang, Haotian Zhang, Renshu Gu, and Jenq-Neng Hwang.
\newblock Exploit the connectivity: {M}ulti-object tracking with
  {T}racklet{N}et.
\newblock In {\em Proceedings of the 27th ACM International Conference on
  Multimedia}, pages 482--490, 2019.

\bibitem{scheidegger2018mono}
Samuel Scheidegger, Joachim Benjaminsson, Emil Rosenberg, Amrit Krishnan, and
  Karl Granstr{\"o}m.
\newblock Mono-camera 3{D} multi-object tracking using deep learning detections
  and {PMBM} filtering.
\newblock In {\em 2018 IEEE Intelligent Vehicles Symposium (IV)}, pages
  433--440. IEEE, 2018.

\bibitem{sharma2018beyond}
Sarthak Sharma, Junaid~Ahmed Ansari, J~Krishna Murthy, and K~Madhava Krishna.
\newblock Beyond {Pixels: Leveraging Geometry and Shape Cues for Online
  Multi-Object T}racking.
\newblock In {\em 2018 IEEE International Conference on Robotics and Automation
  (ICRA)}, pages 3508--3515. IEEE, 2018.

\bibitem{bai2017deep}
Min Bai and Raquel Urtasun.
\newblock Deep {Watershed Transform for Instance Segmentation}.
\newblock In {\em Proceedings of the IEEE/CVF Conference on Computer Vision and
  Pattern Recognition (CVPR)}, pages 5221--5229, 2017.

\bibitem{redmon2016you}
Joseph Redmon, Santosh Divvala, Ross Girshick, and Ali Farhadi.
\newblock You {O}nly {L}ook {O}nce: Unified, real-time object detection.
\newblock In {\em Proceedings of the IEEE/CVF Conference on Computer Vision and
  Pattern Recognition (CVPR)}, pages 779--788, 2016.

\bibitem{xiong2019upsnet}
Yuwen Xiong, Renjie Liao, Hengshuang Zhao, Rui Hu, Min Bai, Ersin Yumer, and
  Raquel Urtasun.
\newblock {UPSN}et: {A} {U}nified {P}anoptic {S}egmentation {N}etwork.
\newblock In {\em Proceedings of the IEEE/CVF Conference on Computer Vision and
  Pattern Recognition (CVPR)}, pages 8818--8826, 2019.

\bibitem{Simonyan15}
Karen Simonyan and Andrew Zisserman.
\newblock Very deep convolutional networks for large-scale image recognition.
\newblock In {\em International Conference on Learning Representations}, 2015.

\bibitem{ioffe2015batch}
Sergey Ioffe and Christian Szegedy.
\newblock Batch {Normalization: Accelerating Deep Network Training by Reducing
  Internal Covariate Shift}.
\newblock In {\em International Conference on Machine Learning}, pages
  448--456, 2015.

\bibitem{kuhn1955hungarian}
Harold~W Kuhn.
\newblock The {H}ungarian method for the assignment problem.
\newblock {\em Naval research logistics quarterly}, 2(1-2):83--97, 1955.

\bibitem{caesar2019nuscenes}
Holger Caesar, Varun Bankiti, Alex~H Lang, Sourabh Vora, Venice~Erin Liong,
  Qiang Xu, Anush Krishnan, Yu~Pan, Giancarlo Baldan, and Oscar Beijbom.
\newblock {n}u{S}cenes: A multimodal dataset for autonomous driving.
\newblock {\em arXiv preprint arXiv:1903.11027}, 2019.

\bibitem{milan2016mot16}
Anton Milan, Laura Leal-Taix{\'e}, Ian Reid, Stefan Roth, and Konrad Schindler.
\newblock {MOT}16: {A benchmark for Multi-Object T}racking.
\newblock {\em arXiv preprint arXiv:1603.00831}, 2016.

\bibitem{paszke2019pytorch}
Adam Paszke, Sam Gross, Francisco Massa, Adam Lerer, James Bradbury, Gregory
  Chanan, Trevor Killeen, Zeming Lin, Natalia Gimelshein, Luca Antiga, et~al.
\newblock Py{T}orch: An imperative style, high-performance deep learning
  library.
\newblock In {\em Advances in Neural Information Processing Systems}, pages
  8024--8035, 2019.

\bibitem{xiang2017posecnn}
Yu~Xiang, Tanner Schmidt, Venkatraman Narayanan, and Dieter Fox.
\newblock Pose{CNN}: {A Convolutional Neural N}etwork for 6{D} {Object Pose
  Estimation in Cluttered S}cenes.
\newblock {\em Robotics: Science and Systems (RSS)}, 2018.

\bibitem{lin2014microsoft}
Tsung-Yi Lin, Michael Maire, Serge Belongie, James Hays, Pietro Perona, Deva
  Ramanan, Piotr Doll{\'a}r, and C~Lawrence Zitnick.
\newblock Microsoft {COCO}: {Common Objects in C}ontext.
\newblock In {\em European Conference on Computer Vision}, pages 740--755.
  Springer, 2014.

\bibitem{neuhold2017mapillary}
Gerhard Neuhold, Tobias Ollmann, Samuel Rota~Bulo, and Peter Kontschieder.
\newblock The {M}apillary vistas dataset for semantic understanding of street
  scenes.
\newblock In {\em Proceedings of the IEEE International Conference on Computer
  Vision}, pages 4990--4999, 2017.

\bibitem{li2018deepim}
Yi~Li, Gu~Wang, Xiangyang Ji, Yu~Xiang, and Dieter Fox.
\newblock Deep{IM}: {Deep Iterative M}atching for 6{D} {Pose E}stimation.
\newblock In {\em Proceedings of the European Conference on Computer Vision
  (ECCV)}, pages 683--698, 2018.

\bibitem{oberweger2018making}
Markus Oberweger, Mahdi Rad, and Vincent Lepetit.
\newblock Making deep heatmaps robust to partial occlusions for 3{D} object
  pose estimation.
\newblock In {\em Proceedings of the European Conference on Computer Vision
  (ECCV)}, pages 119--134, 2018.

\bibitem{peng2019pvnet}
Sida Peng, Yuan Liu, Qixing Huang, Xiaowei Zhou, and Hujun Bao.
\newblock Pvnet: Pixel-wise voting network for 6{D}o{F} pose estimation.
\newblock In {\em Proceedings of the IEEE Conference on Computer Vision and
  Pattern Recognition}, pages 4561--4570, 2019.

\bibitem{zakharov2019dpod}
Sergey Zakharov, Ivan Shugurov, and Slobodan Ilic.
\newblock {DPOD}: 6{D} {Pose Object Detector and R}efiner.
\newblock In {\em Proceedings of the IEEE International Conference on Computer
  Vision}, pages 1941--1950, 2019.

\bibitem{kendall2015posenet}
Alex Kendall, Matthew Grimes, and Roberto Cipolla.
\newblock Pose{Net: A Convolutional Network for R}eal-time 6-{D}o{F} {Camera
  R}elocalization.
\newblock In {\em Proceedings of the IEEE International Conference on Computer
  Vision}, pages 2938--2946, 2015.

\bibitem{long2015fully}
Jonathan Long, Evan Shelhamer, and Trevor Darrell.
\newblock Fully {Convolutional Networks for Semantic S}egmentation.
\newblock In {\em Proceedings of the IEEE/CVF Conference on Computer Vision and
  Pattern Recognition (CVPR)}, pages 3431--3440, 2015.

\bibitem{he2016deep}
Kaiming He, Xiangyu Zhang, Shaoqing Ren, and Jian Sun.
\newblock Deep residual learning for image recognition.
\newblock In {\em Proceedings of the IEEE/CVF Conference on Computer Vision and
  Pattern Recognition (CVPR)}, pages 770--778, 2016.

\bibitem{bernardin2008evaluating}
Keni Bernardin and Rainer Stiefelhagen.
\newblock Evaluating {Multiple Object Tracking P}erformance: the {CLEAR} {MOT}
  metrics.
\newblock {\em EURASIP Journal on Image and Video Processing}, 2008:1--10,
  2008.

\end{thebibliography}


\end{document}